\documentclass[conference]{IEEEtran}
\IEEEoverridecommandlockouts
\usepackage{cite}
\usepackage{amsmath,amssymb,amsfonts}
\usepackage{algorithmic}
\usepackage{graphicx}
\usepackage{textcomp}
\usepackage{multirow}
\usepackage{multicol}
\usepackage{mwe}
\usepackage{marginnote}
\usepackage{soul}
\usepackage{xargs}   
\usepackage{graphicx}
\usepackage{subcaption}
\usepackage{flushend} 
\usepackage[pdftex,dvipsnames]{xcolor}  
\usepackage{array}
\usepackage{algorithm}
\usepackage{algorithmic}
\usepackage[compact]{titlesec}
\usepackage{float}
\usepackage{titlesec}

\usepackage{enumitem}
\setlist[itemize]{nosep}
\usepackage[acronym]{glossaries}
\glsdisablehyper
\makeglossaries   
\newacronym{hpc}{HPC}{High Performance Computing}
\newacronym{ml}{ML}{Machine Learning}
\newacronym{ping}{PinG}{Performance in a Graph}
\newacronym{mse}{MSE}{Mean Square Error}
\newacronym{dnn}{DNN}{Deep Neural Network}
\newacronym{gnn}{GNN}{Graph Neural Network}
\newacronym{gcn}{GCN}{Graph Convolutional Network}
\newacronym{gat}{GAT}{Graph Attention Layer}
\newacronym{sgc}{SGC}{Single-Graph Construction Approach}
\newacronym{bgc}{BGC}{Batched-Graph Construction Approach}
\newacronym{dae}{DAE}{Denoising Autoencoder}
\newacronym{ssgnn}{SSGNN}{Self-supervised Graph Neural Network}
\newacronym{ssbgnn}{SSBGNN}{Self-supervised Batched Graph Neural Network}

\newcommand{\tzi}[1]{\textcolor{black}{#1}}
\usepackage[colorinlistoftodos,prependcaption]{todonotes}

\def\BibTeX{{\rm B\kern-.05em{\sc i\kern-.025em b}\kern-.08em
    T\kern-.1667em\lower.7ex\hbox{E}\kern-.125emX}}
\begin{document}

\title{Novel Representation Learning Technique using Graphs for Performance Analytics
}

\author{\IEEEauthorblockN{Tarek Ramadan}
\IEEEauthorblockA{\textit{Oracle Corporation}\\
tarek.ramadan@oracle.com}
\and
\IEEEauthorblockN{Ankur Lahiry}
\IEEEauthorblockA{\textit{Texas State University}\\
vty8@txstate.edu}
\and
\IEEEauthorblockN{Tanzima Z. Islam}
\IEEEauthorblockA{\textit{Texas State University}\\
tanzima@txstate.edu}
}

\maketitle

\begin{abstract}
The performance analytics domain in \gls{hpc} uses tabular data to solve regression problems, such as predicting the execution time. Existing \gls{ml} techniques leverage the correlations among features given tabular datasets, not leveraging the relationships between samples directly. 
Moreover, since high-quality embeddings from raw features improve the fidelity of the downstream predictive models, existing methods rely on extensive feature engineering and pre-processing steps, costing time and manual effort. To fill these two gaps, we propose a novel idea of transforming tabular performance data into graphs to leverage the advancement of Graph Neural Network-based (GNN) techniques in capturing complex relationships between features and samples. In contrast to other ML application domains, such as social networks, the graph is not given; instead, we need to build it. To address this gap, we propose graph-building methods where nodes represent samples, and the edges are automatically inferred iteratively based on the similarity between the features in the samples. We evaluate the effectiveness of the generated embeddings from GNNs based on how well they make even a simple feed-forward neural network perform for regression tasks compared to other state-of-the-art representation learning techniques. Our evaluation demonstrates that even with up to 25\% random missing values for each dataset, our method outperforms commonly used graph and \gls{dnn}-based approaches and achieves up to 61.67\% \& 78.56\%  improvement in MSE loss over the DNN baseline respectively for HPC dataset and Machine Learning Datasets.
\end{abstract}


\section{Introduction}
\label{sec:intro}
Simulations leverage \gls{hpc} systems to evaluate numerous what-if scenarios of otherwise experimentally intractable phenomena. HPC configurations (e.g., number of threads, nodes, power cap, thread binding) significantly impact system utilization and execution time~\cite{panwar2019quantifying,patki2019performance,islam2016a,beckingsale2017apollo}. Estimating how long an application will run based on the configurations \textit{before submitting the job} is a critical problem in \gls{hpc} since scheduling systems use this estimation as a hard cut-off for aborting the application and scheduling a new job~\cite{Tang1,Fan:UnderestimationJobRuntime}. Once an application gets evicted from an \gls{hpc} system, the scheduler adds it at the back of the job queue. At a time, thousands of jobs can be waiting in the queue, meaning a user may need to wait for their job to be scheduled again, delaying the overall turnaround time of science. 
On the other hand, over-predicting the runtime causes the system to be idle once the application finishes leaving billion-dollar systems under-utilized. Hence, over and under predictions of the runtime is undesirable. 

\begin{figure}[t]
\centering
  \includegraphics[width=\columnwidth]{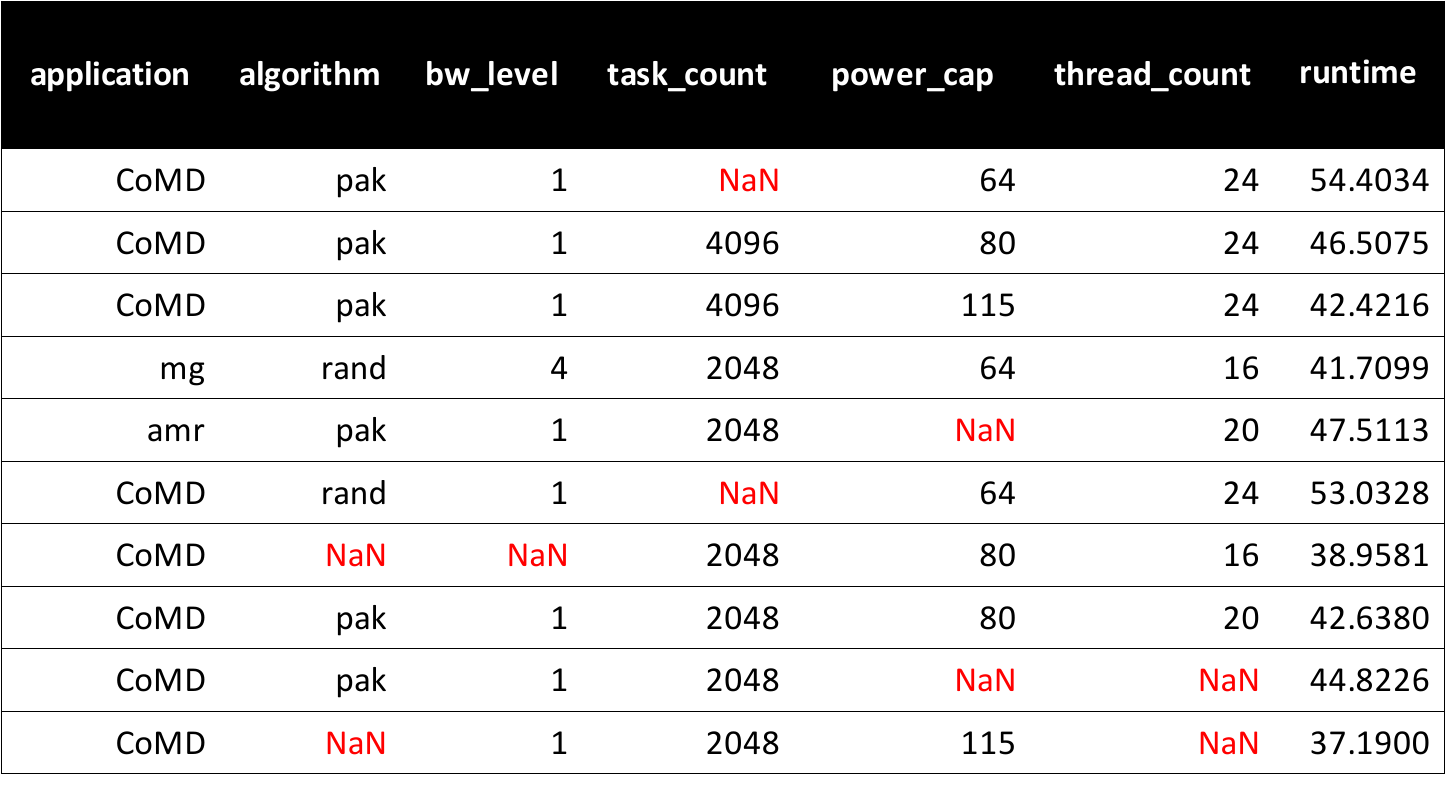}
\caption{A sample performance dataset. It contains numerical and categorical features in the first six columns and the target value in the last (Runtime). Note that values colored in red (NaN) are missing values. Streaming \gls{hpc} performance monitoring systems often miss recording feature values or cannot measure the runtime until the application ends, resulting in missing feature values and unlabeled samples.
}
  \label{fig:Tabular Data}
  \vspace{-0.2in}
\end{figure}

The performance analytics research area in \gls{hpc} leverages \gls{ml} techniques to enable precise runtime prediction. Since the success of downstream \gls{ml} tasks depends on data representation~\cite{jonschkowski2014state,bengio2012unsupervised}, in this paper, we investigate the problem of extracting the most meaningful information from seemingly unrelated user inputs (i.e., configurations and algorithms) to preserve the correlations between input features and target performance metrics (e.g., runtime). 
Most \gls{hpc} performance datasets are organized in tabular format. 
Figure~\ref{fig:Tabular Data} shows an example of a tabular dataset collected for \gls{hpc} performance analytics, where the columns contain features including \texttt{application}, \texttt{algorithm}, \texttt{power\_cap}, and the rows contain samples. The target column is denoted as \texttt{runtime}, which describes how long an application takes when being run with a specific \texttt{algorithm}, \texttt{bandwidth level}, \texttt{task count}, \texttt{power cap}, and \texttt{thread count}.
While the existing performance analytics research captures user inputs and the respective application performance in tabular format, this format only allows most downstream \gls{ml} models to exploit relationships across the features within each sample but not across all samples directly.

Finding an effective representation learning technique for tabular data is an active research area. Tree-based models such as XGBoost~\cite{chen2015xgboost} have achieved encouraging enhancement in improving the fidelity of regression problems for real-world applications~\cite{ding2020computational,osman2021extreme,kiangala2021effective}. \textbf{Since missing measurements of hardware features is common in \gls{hpc} (represented as NaN values for certain features), 
existing supervised algorithms perform poorly or become inapplicable~\cite{emmanuel2021survey} in predicting the execution time using NaN-filled fetures}. 
\tzi{Moreover, performance samples in real-time monitoring systems arrive in a stream, which makes tabular methods rebuild models from scratch every time a new sample appears.}
To overcome these issues, existing literature has proposed an Attentive Interpretable Tabular Learning (AITL) technique such as the TabNet~\cite{arik2021tabnet} model. Nonetheless, the TabNet model also uses deep learning techniques to leverage the correlations among features within each sample 
while leaving relationships among the samples largely unexploited. 
To overcome this shortcoming, the literature uses extensive feature engineering and pre-processing steps, which come at a high cost and requires a human in the loop.

To address the gaps mentioned above, \textbf{we propose to investigate a novel approach of transforming tabular data into a graph data structure to model the correlations among features and samples explicitly.} Specifically, \textit{we hypothesize that using a graph structure to describe a performance dataset (spanning many samples) 
makes similarities between samples and features explicit, thus constructing the idea of neighborhoods.} 
The rationale for organizing samples based on similarity is that performance samples with similar feature values, e.g., \texttt{number of threads} = 4 and 6, are likely to result in similar runtimes. In contrast, samples with vastly different features, e.g., \texttt{number of threads} = 4 and 16, are likely to result in widely different runtimes. 

However, unlike social networks or disease propagation domains, a performance graph is not given. 
We overcome this gap by proposing two different approaches for constructing a performance graph called \gls{ping} and evaluating them based on how they impact the effectiveness of the downstream analysis techniques. Another motivation of the \gls{ping} formulation is that for streaming performance analytics systems, where labeled data is scarce, and it is too expensive to rebuild models when new samples arrive, new unlabeled samples can be placed in the existing dataset based on their inherent neighborhood similarity. \textbf{To our knowledge, the idea of transforming tabular performance data into graphs to improve regression models' effectiveness has never been explored in the HPC performance analytics domain.}

To summarize the contributions of this paper, we:
\begin{itemize}
\item Transform tabular performance data into a graph data structure (\gls{ping}) to capture relationships between samples and features.
\item Build the \gls{ping} structure using two methods (since the performance graph is not given) and explore their effectiveness in improving the performance of downstream regression tasks. 
\item Develop a novel representation learning technique that can automatically refine the edges between samples (nodes in \gls{ping}) based on feature and sample similarities using self-supervised Graph Neural Networks (GNNs). 
\item Develop an end-to-end framework that implements the proposed data transformation and representation learning techniques to build effective embeddings for explainable downstream models. 
\end{itemize}

We evaluate the effectiveness of the generated embeddings from GNNs based on how well they make even a simple feed-forward neural network perform for regression tasks compared to other state-of-the-art representation learning techniques. \textbf{Since the characteristics of problems in \gls{hpc} are unique, e.g., regression tasks, missing labels, missing measurements, streaming data, scarcity of training samples, this paper focuses solely on the applicability of our proposed approach in the context of the \gls{hpc} domain.} Thus, we use ten \gls{hpc} performance datasets collected on three supercomputing facilities. To study the applicability of our proposed design on ML datasets, we also present evaluation using three ML benchmarks. 
Our evaluation demonstrates that even with up to 25\% random missing values for each dataset, our method outperforms commonly used graph and \gls{dnn}-based approaches. Specifically, our method achieves up to 61.67\% \& 78.56\% improvement in \gls{mse} loss over \gls{dnn} for the HPC and ML datasets, respectively.

\section{Design}
\label{sec:design}
\begin{figure}[t]
\centering
\includegraphics[width=\columnwidth]{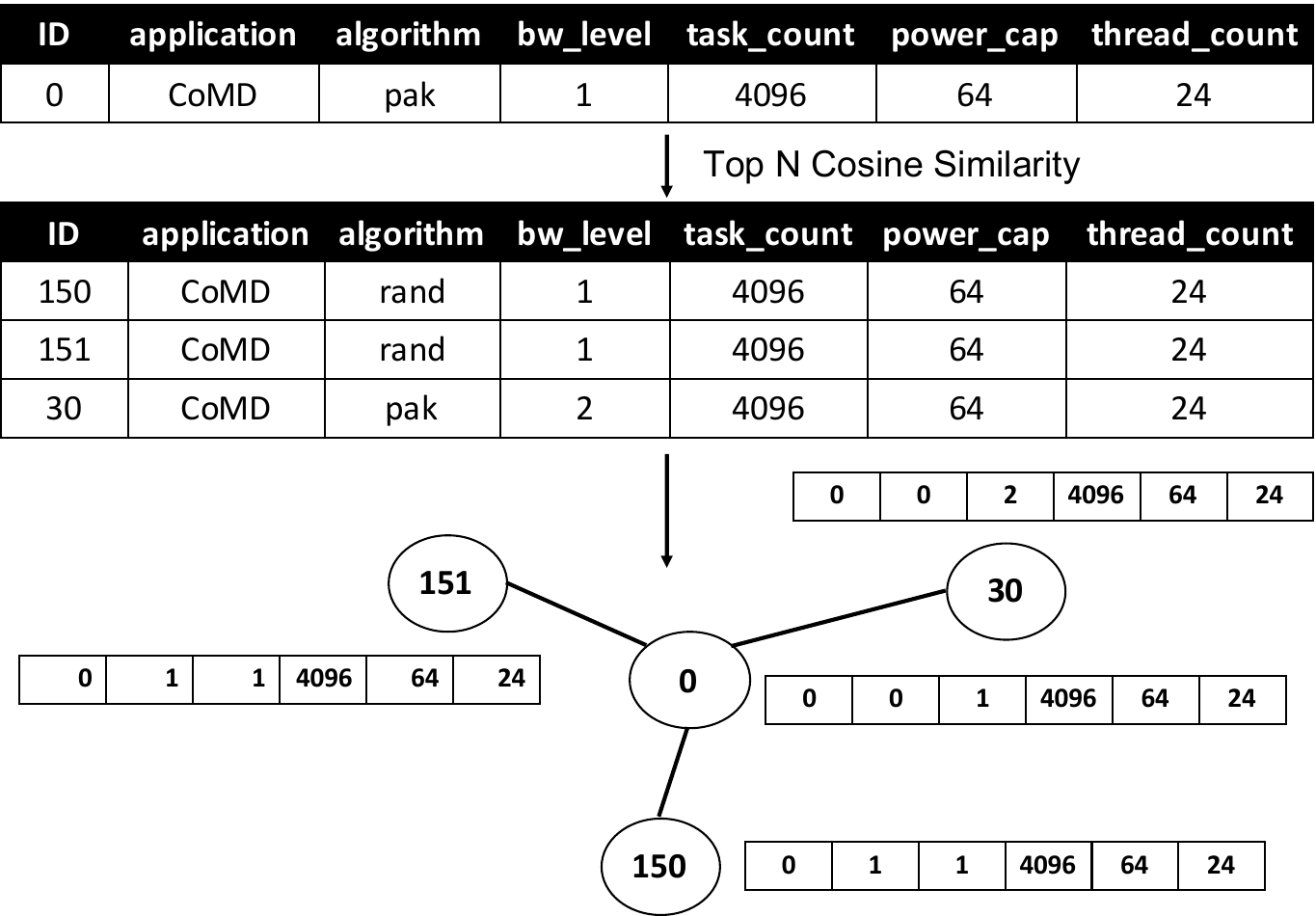}
\caption{Illustration of the Single-graph construction approach using $N$ Cosine similarity. Here, $N$ is a user-defined parameter that determines the density of the performance graph.}
\label{fig:SimEx}
\vspace{-0.2in}
\end{figure}
\begin{figure}[t]
\centering
\includegraphics[width=\columnwidth]{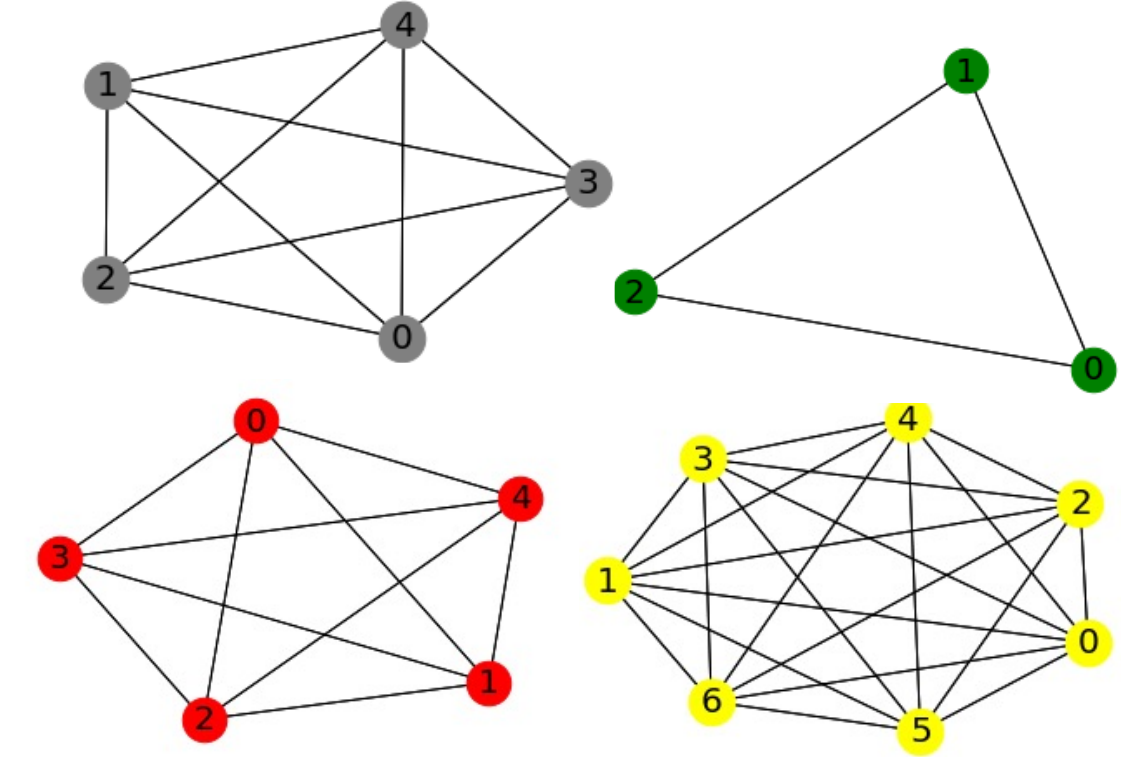}
\caption{Outcome of the batched-graph construction approach.}
\label{fig:batchedgraph}
\vspace{-0.2in}
\end{figure}
We hypothesize that holistically capturing the relationships across samples and parameters enabled by the graph formulation of tabular data can improve the effectiveness of the downstream \gls{ml} tasks. 

\subsection{Rationale for \gls{ping}}
The rationale for organizing data as a graph compared to the state-of-the-art dictionary-based approach is that a graph structure can inherently describe relationships among features and samples, allowing downstream ML tasks to capture information from relevant neighbors by exploiting these interrelations. However, unlike other research domains where the graph structure is explicitly provided, the graph from \gls{hpc} data needs to be constructed. While data samples can directly map to nodes, edges between samples need to be defined explicitly or inferred automatically. On the other hand, the dictionary-based approach only leverages feature extraction to generate the representation vector, whereas the graph structure leverages both feature extraction and the relationship between the samples in generating the representation vector.

The graph design adopts the architecture of an undirected weighted graph $G = (V, E, A)$ to represent a performance dataset, where each node $V_i \in \mathrm{V}$ denotes a sample, and $edge \in \mathrm{E}$ indicates a relationship with another measurement. $A \in \mathrm{R^{NxN}}$ is an adjacency matrix that specifies the weights on the edges, where $A_{i,j}$ corresponds to the edge weight between nodes $V_i$ and $V_j$ that can be calculated based on their similarities. For data that belongs to the Euclidean space, $A_{ij}$  can be computed using a distance metric between $V_i$ and $V_j$. However, for non-Euclidean or dense data such as a tree, building $A_{ij}$ using Euclidean distance is not meaningful.
Hence, instead of using explicit distance measures to construct the final graph, we propose to \textit{first,} construct a fully connected initial graph using measurements as node features and explicit distance measures such as cosine similarity to calculate the initial edge weights, and \textit{then,} iteratively 
refine the node and edge features through an automated graph edge inference method using self-supervised learning.

\subsection{Initial \gls{ping} Construction}
\label{sec:construction}

This section presents two approaches for generating the initial \gls{ping} graph from the tabular data. 

\noindent
\textbf{\gls{sgc}: } In this approach, we propose to build a single graph for the whole dataset where each sample represents a node. 
The initial edges in \gls{ping} are computed using 
the Cosine similarity algorithm 
by measuring the cosine angle between the current sample and all other samples. A high cosine similarity indicates that the angle between the two samples is small; therefore, the two samples are similar to each other. Using this technique, we calculate the cosine similarity for each sample with all different samples in the dataset and then choose the top $N$ similar samples. Here, $N$ is a hyper-parameter that users can adjust for better model performance or specify based on memory constraints. Figure~\ref{fig:SimEx} demonstrates the cosine-similarity-based neighbor selection for a given node where $N$ = 3. This step can be parallelized using a data-parallel ML technique that we will pursue in the future.
After building the edges between similar samples, the edge weights are initialized to $1$ and passed to a GNN model 
using self-supervised learning. The model assigns higher edge weight to similar data and lower edge weight to dissimilar data, thus explicitly encoding the similarity between samples and their features. Algorithm~\ref{alg:scg} shows the algorithmic steps of this process.
\begin{algorithm}
\caption{Single Graph Construction Approach}
\label{alg:scg}
\begin{algorithmic}[1]
\STATE \textbf{Input:} $x$ is a tabular dataset., $n$ minimum number of neighbours
\STATE \textbf{Output:} $graph$, a NetworkX graph generated from the tabular dataset
\STATE $distance \gets$  $cosine\_distance(dataset)$
\STATE declare empty arrays of $sources$ and $destinations$
\FORALL{$index, distance \in distances$}  
    \STATE $source \gets$ [1, N + 1]
    \STATE append $source$ in $sources$ 
    \STATE $similars \gets$ $sort(distances)$
    \STATE $destination \gets$ $similars.top(N+1)$
    \STATE append $destination$ in $destinations$
\ENDFOR
\STATE $edges \gets$ zipped(sources, destinations)
\STATE $graph \gets$ an empty NetworkX graph
\STATE create [1...N+1] nodes in the $graph$ 
\STATE add edges from the $edges$ list
\STATE add node features and edge features in the $graph$
\STATE remove self-loop from the $graph$
\RETURN $graph$
\end{algorithmic}
\end{algorithm}

\noindent
\textbf{\gls{bgc}: } In this approach, we propose to generate clusters of samples using unsupervised learning to find groups within the data based on similarity to build $M$ graphs, where $M$ is the number of clusters. This approach is memory efficient since it uses batches of samples instead of all samples for building each sub-graph centered around each sample 
and leverage the averaging effect. 
The \gls{bgc} method selects different cluster sizes to create different-sized batches, as illustrated in Figure~\ref{fig:batchedgraph}, thus the cluster size is a hyperparameter to our algorithm. To further optimize this approach, we can reduce the number of edges by connecting the top $N$ similar neighbors to each node, as explained in the first approach. We demonstrate the process in Algorithm~\ref{alg:bgc}.
In this paper, we choose Hierarchical Density-based Spatial Clustering of Applications with Noise (HDBscan)~\cite{mcinnes2017hdbscan} due to its parameter flexibility. 
HDBscan encodes the input samples into transformed space according to their density, then builds a minimum spanning tree (MST) 
to find the cluster hierarchy, and finally extracts the generated cluster.

\begin{algorithm}
\caption{Batched-Graph Construction Approach}
\label{alg:bgc}
\begin{algorithmic}[1]
\STATE \textbf{Input:} $x$ is a tabular dataset., $N$ minimum number of neighbours
\STATE \textbf{Output:} $graphs$, batched generation of graphs
\STATE $graphs \gets$ empty array
\STATE $clusters \gets$ a group of clusters from the sample
\FORALL{$group \in clusters$}
    \STATE $distances \gets$ $cosine\_distance(group)$ 
    \STATE declare empty arrays of $sources$ and $destinations$
    \FORALL{$index, distance \in distances$}  
    \STATE $source \gets$ [1, N + 1]
    \STATE append $source$ in $sources$ 
    \STATE $similars \gets$ $sort(distances)$ 
    \STATE $destination \gets$ $similars.top(N+1)$
    \STATE append $destination$ in destinations
    \STATE $edges \gets$ zipped(sources, destinations)
    \STATE $graph \gets$ an empty NetworkX graph
    \STATE create [1...N+1] nodes in the $graph$ 
    \STATE add edges from the $edges$ list
    \STATE add node features and edge features in the $graph$
    \STATE remove self-loop from the graph
    \STATE append $graph$ in $graphs$
\ENDFOR
\ENDFOR
\RETURN $graphs$
\end{algorithmic}
\vspace{-0.05in}
\end{algorithm}
\subsection{Self-Supervised Learning}
Since a large number of streaming performance samples in \gls{hpc} are unlabeled, supervised learning is not ideal.
In contrast, Self-supervised learning (SSL) algorithms can recognize and understand patterns from unlabeled samples. The main idea of SSL is to automatically generate a representation of the input data and refine it using feedback from the model, showing promising results in Natural Language Processing (NLP)~\cite{radford2019language,devlin2018bert}. 
Hence, in this paper, we propose leveraging self-supervised learning to develop an automated edge-inference-based method that optimizes the \gls{ping} edge weights.

\subsection{Putting It Altogether for Representation Learning}
\label{sec:rep-learn}
\begin{figure}[t]
\centering
  \includegraphics[width=\columnwidth]{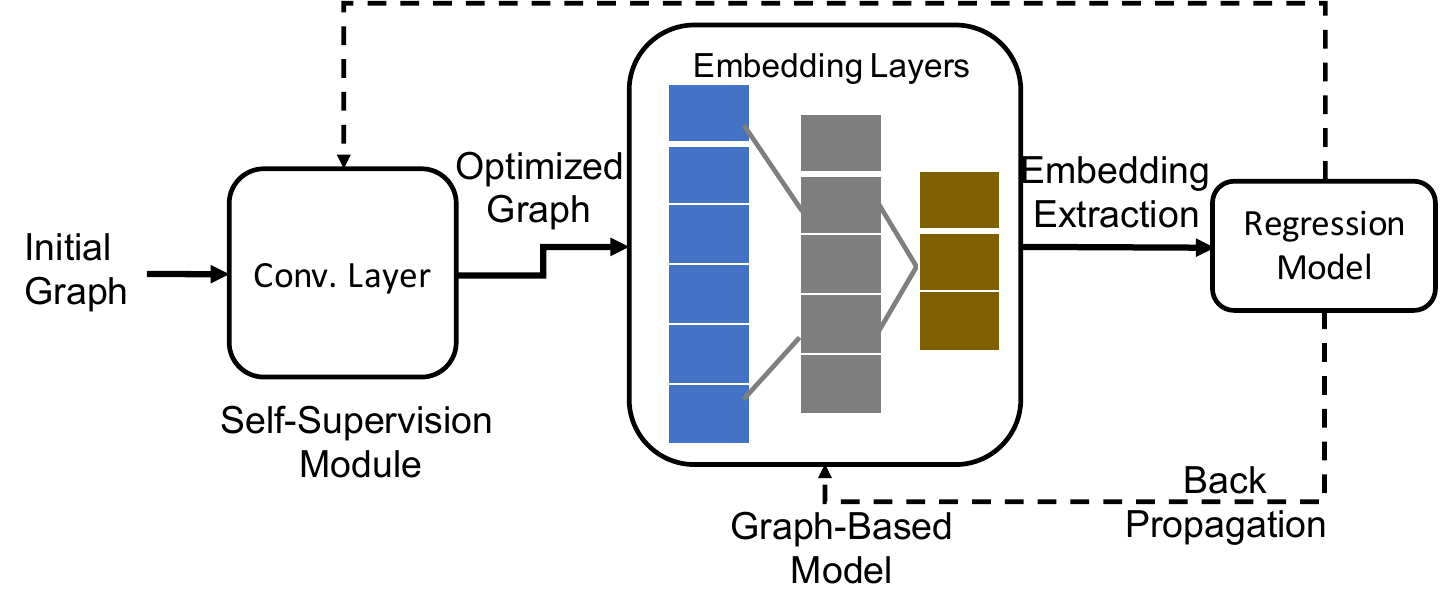}
\caption{The end-to-end pipeline of learning the optimized graph from initially constructed \gls{ping} formulation. Input graph and its edge weights are fed to a \gls{gnn} model with self-supervision for constructing an embedding for the entire dataset. Here, solid lines represent forward propagation, and dotted lines represent backpropagation to enhance the edge weights.}
  \label{fig:Framework1}
  \vspace{-0.2in}
\end{figure}

\gls{gnn}-based representation learning techniques~\cite{wu2021learning} can represent nodes, edges, and sub-graphs in low-dimensional vectors. These techniques capture complex and non-linear relationships among samples and features across the whole dataset to build the embeddings. 
Figure \ref{fig:Framework1} presents our proposed end-to-end representation learning pipeline. After the initial graph is constructed using one of the \gls{sgc} or \gls{bgc} methods presented in Section~\ref{sec:construction}, the initial graph is passed to a convolution layer for creating self-supervision and then passed to a graph-based learning model for building embeddings. The embeddings are then extracted from the second-to-last layer and passed to a feed-forward neural network to calculate \gls{mse}. The error then propagates back to the self-supervision and graph-based models to provide feedback and fine-tune.

Specifically, in this work, we propose to use \gls{gnn}~\cite{scarselli2008graph} to learn the embeddings from the constructed graphs. 
The main advantage of the GNN architecture is that the learning equation leverages edges connected between a node to its neighbors to compute that node's vector representation. This operation is known as message passing. One of the most recent \gls{gnn} architectures is GraphSAGE~\cite{hamilton2017inductive}, which learns each node's vector representation inductively by allowing complex aggregation functions for message passing. Thus, message passing makes \gls{gnn} generalize to unseen samples during the deployment of the model by representing the unseen sample to the neighboring nodes. The GraphSAGE model generates a vector representation for each node and then applies semi-supervised node regression to produce a regression value for each node. We refer to the representation learning pipeline that uses \gls{sgc} as the input as \gls{ssgnn}, and that using \gls{bgc} as \gls{ssbgnn}. In Section~\ref{sec:results}, we compare the performance of using GraphSAGE as the main driver of \gls{ssgnn} and \gls{ssbgnn} with that of other methods such as \gls{gcn} and \gls{gat}. \textbf{Our extensive experiments show that the choice of the driver method can impact accuracy of our proposed method and is dataset dependent.}

\section{Implementation}
\label{sec:implementation}
We implement the proposed representation learning pipeline using the Networkx (NX)~\cite{hagberg2008exploring} and Deep Graph Library (DGL)~\cite{wang2019deep} in Python with the PyTorch back end.

\subsection{Pre-processing}
Since the core of our approach is to find a better representation of the input data automatically, we used minimal processing, precisely two pre-processing techniques: standardization and imputation of missing values.

\noindent
\textbf{Standardization: }To standardize the input data into a common scale, our pipeline transforms the data using the standard \texttt{scaler} algorithm provided by sklearn~\cite{pedregosa2011scikit}. The standard \texttt{scaler} algorithm removes the mean and scales the data based on the unit variance so that the scaling of each feature is independent of the other features; however, eventually, they are all scaled in the same way and have equal weights.

\noindent
\textbf{Imputation: }To address the missing values in the input data, our pipeline uses the \texttt{Simpler Imputer} function provided by sklearn. The \texttt{Simpler Imputer} function finds the missing values and replaces all missing values in the column based on a selected strategy. We used the feature column's mean to replace any missing values.

\subsection{Classifier Optimization}

\noindent

\noindent
\textbf{Drop-out Layer:} We randomly drop some of the edges to the subsequent layers' neurons during the training phase. The drop-out probability is a hyperparameter. This technique is only applied to the input and hidden layers of the model during training and does not affect the testing set.

\noindent
\textbf{Optimizer:} The loss gets calculated based on the loss function in every epoch, and the optimizer tweaks the model parameters to minimize the loss. 
This work uses the Adam optimizer~\cite{kingma2014adam}. 

\noindent
\textbf{Activation Function:} 
The pipeline uses an activation function in each neuron node in the neural network. Based on multiple experiments, we choose Rectified Linear Unit (ReLU) as the activation function. 

\noindent
\textbf{Removing self loops:}
A node's closest neighbor when computing cosine distance is the node itself, which introduces self-loops in the graph. So, while choosing the top $N$ neighbors, we consider the top $N + 1$ neighbors first instead (line 12 in Algorithm~\ref{alg:bgc} \& line 9 in Algorithm~\ref{alg:scg}) and then remove those self-loops from the induced graph. 

\section{Experimental Setup}
\label{sec:setup}

\subsection{System}
\begin{table}[t]
\footnotesize
\caption{Regression dataset descriptions. ``\#S", ``\#F", and ``Std.Dev." refer to the number of samples, number of features, and standard deviation in the target values of the corresponding dataset.}
\label{table:Datasets description}
\renewcommand{\arraystretch}{1.2}
\centering
\begin{tabular}{|p{1.3cm}|p{0.6cm}|p{0.4cm}|p{1cm}|p{0.9cm}|p{0.6cm}|p{0.9cm}|}
\hline
Dataset	&\#S	&\#F	&Domain	&Target	&Mean &Std.Dev \\ \hline
Catalyst\cite{patki2019performance}	&3992	&10	&HPC	&Runtime	&36.88&17.38 \\ \hline
Cab\cite{tanzima_2019_3403038}	&320	&148	&HPC	&Runtime	&21.04&15.65 \\ \hline
Vulcan\cite{tanzima_2019_3403038}	&321	&172	&HPC	&Runtime	&23.51 &29.43 \\ \hline
MiniAMR \cite{miniAMR}	&301	&9	&HPC	&Runtime	&34.68 &18.06 \\ \hline
CG\cite{naspar}	&961	&9	&HPC	&Runtime	&31.48 &13.75 \\ \hline
FT\cite{naspar} &554	&9	&HPC	&Runtime	&34.15 & 16.90\\ \hline
LU\cite{naspar} &554	&9	&HPC	&Runtime	&43.10 &22.09 \\ \hline
MG\cite{naspar}	&513	&9	&HPC	&Runtime	&48.65 & 23.98\\ \hline
CoMD\cite{comd}	&553	&9	&HPC	&Runtime	&41.61 &4.71 \\ \hline
Kripke\cite{kripke}	&554	&9	&HPC	&Runtime	&28.29 &4.49\\ \hline
Airfoil Self-Noise\cite{misc_airfoil_self-noise_291} & 1503 & 6 &ML &SSPL &124.84 &6.90  \\ \hline
IoT Telemetry \cite{iot_telemetry} & 1000 & 6 &ML &Temp &22.36 &2.56 \\ \hline
Beijing AQI \cite{misc_beijing_multi-site_air-quality_data_501} & 1000 & 15 &ML &PT08.S4 (NO2) &9.76  &43.69 \\ \hline
\end{tabular}
\vspace{-0.2in}
\label{tab:datasets}
\end{table}

We run our experiments in parallel using the following systems:
Penguin Computing On Demand (POD), Google Colab, Texas Advanced Computing Center (TACC). The POD system consists of 96 AMD EPYC CPUs. Each CPU has 8GB of RAM based on the 64-bit x86 architecture. 
We also run our evaluations on Google Colab using a machine with Intel(R) Xeon(R) CPU with 26.75GB RAM and one NVIDIA Tesla P100 GPU with 16GB memory. Since this work involves running various models simultaneously, we leverage supercomputers at the TACC center to run them in parallel. 
\subsection{Dataset Description}
To thoroughly evaluate the benefit of our proposed graph construction and representation learning technique, we use 10 HPC applications and three ML benchmarks in this work. The Vulcan and Cab datasets contain performance data collected from the XSBench~\cite{tramm2014xsbench} and the OpenMC~\cite{romano2013openmc} applications on an IBM BlueGene Q and an Intel Sandy Bridge systems, respectively. The Catalyst dataset combines the performance samples from MiniAMR\cite{miniAMR}, CG\cite{naspar}, FT\cite{naspar}, LU\cite{naspar}, MG\cite{naspar}, CoMD\cite{comd}, Kripke\cite{kripke} miniapps. 
Table~\ref{tab:datasets} summarizes the size, the number of features, and the target variables for each dataset. 

\subsection{Baselines}
In this paper, we compare our proposed GNN-based method to two graph-based representation learning techniques---\gls{gcn}~\cite{haykin2004comprehensive} and \gls{gat}~\cite{velivckovic2017graph}, and four non-graph-based methods commonly used for tabular data---DNN~\cite{schmidhuber2015deep}, XGBoost~\cite{chen2015xgboost}, LightGBM~\cite{ke2017lightgbm}, and \gls{dae}~\cite{vincent2010stacked}. The graph-based methods use the same graph construction approach as the \gls{ssgnn} approach. 
Table~\ref{table:Hyperparameters} represents the hyperparameters for each model and the ranges of their values used during hyperparameter tuning. We leverage the Optuna framework~\cite{akiba2019optuna} to find the best hyperparameters for each model automatically.

\subsection{Performance Metrics}
Since all datasets in HPC performance analytics contain continuous target values, we evaluate the performance of regression tasks by comparing the predicted runtime with the actual one. We use the \gls{mse} loss that calculates the average squares of the error between the predicted and the actual labels. Using different baseline and graph-based methods, we generate embeddings and evaluate their effectiveness by assessing how well a simple linear regression model performs when fed with the embeddings. 
Since this experiment aims to attribute any benefit to having superior embeddings, we keep the prediction task's model simple. 


\begin{table}[t]
\footnotesize
\caption{Hyper-parameters for different models.}
\label{table:Hyperparameters}
\centering
\begin{tabular}{|p{1in}|p{1in}|p{1in}|}
\hline
Model & Hyper-parameters & Search range \\ \hline
\multirow{2}{*}{\parbox{1in}{DNN, XGBoost, LightGBM, DAE, GAT, GCN, SSGNN, and SSBGNN}} 
    &Hidden Layer Dimension &$[25, 600]$\\\cline{2-3}
    & Learning Rate	& $[1e^{-5}, 1e^{-1}]$\\ \cline{2-3}
&Dropout	&[0,1]\\ \cline{2-3}
&Optimizers	 &[Adam, Adagrad, Adadelta, SGD, RMSprop]\\ \cline{2-3}
&Activation function	 &[relu, elu,  leaky\_relu]\\ \hline

\multirow{2}{*}{\parbox{1in}{GAT, GCN, SSGNN, and SSBGNN}} 

   & Hidden Layer Aggregation	&[pool,mean,gcn]\\ \cline{2-3}
&Self-Supervision Conv. Layer activation function&[relu, elu, leaky\_relu, sigmoid, tanh]\\ \cline{2-3}
&Number of Attention Heads
(GAT model only)	&$[1,2]$\\ \hline
\end{tabular}
  \vspace{-0.1in}
\end{table}

\section{Results}
\label{sec:results}
\begin{figure}[t]
    \centering
    \includegraphics[width=\columnwidth]{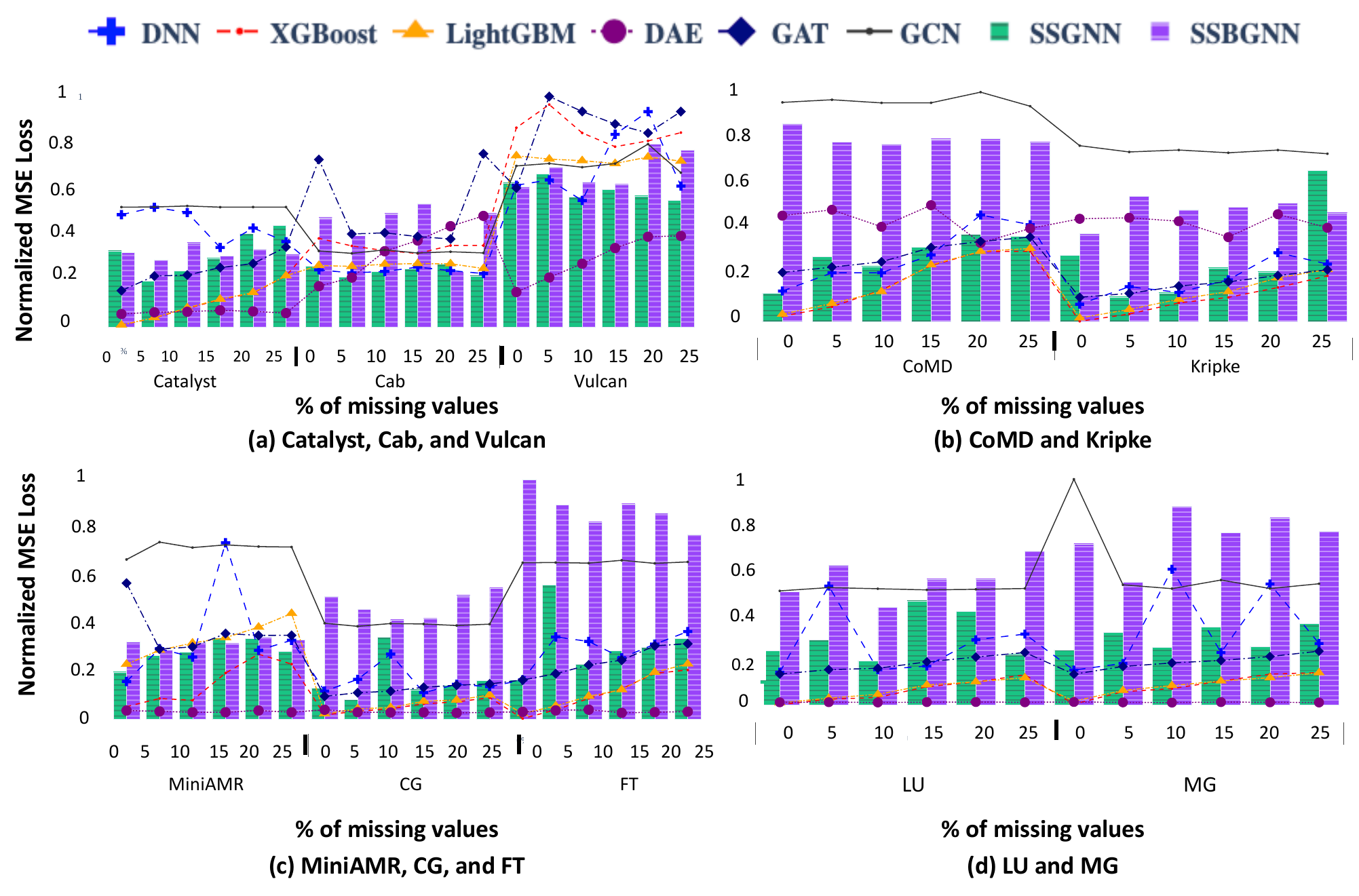}
    \caption{Normalized MSE Loss for HPC dataset The X-axis shows the \% of missing feature values and the Y-axis shows the normalized MSE value for different methods.}
    \label{fig:hpc_mse}
\end{figure}

In all experiments, we use 5-fold cross-validation by dividing the data into five train/test splits, generating five different results, and calculating the average. 
\textbf{We ensure that the train and test sets are disjoint.} Table~\ref{tab:datasets} shows all thirteen datasets from HPC and ML domains and Table~\ref{table:Hyperparameters} shows the hyperparameters for the models used. 

\subsection{Impact of Graph Representation}

\begin{table}[t]
\footnotesize
\caption{Summary of improvements using our proposed methods compared to the baselines across datasets.}
\label{tab:result_table}
\centering
\begin{tabular}{|p{0.2in}|p{0.45in}|p{0.45in}|p{0.45in}|p{0.7in}|}
\hline
Type & \# of Exps. & Methods & Best \% Improvement & \#Exps. where any of Our Methods Outperform  \\  \hline
\multirow{2}{*}{HPC} &
\multirow{2}{*}{60}
    &DNN & 61.67 & 32\\ \cline{3-5}
    &&XGBoost & 38.19 &  12\\\cline{3-5}
    &&LightGBM &  36.03 & 18 \\ \cline{3-5}
    &&DAE & 76.04 &14\\ \cline{3-5}
    &&GAT & 70.55 & 26 \\\cline{3-5}
    &&GCN & 86.96 & 60 \\\hline

\multirow{2}{*}{ML}&
\multirow{2}{*}{18}
    &DNN & 78.56 & 8\\\cline{3-5}
    &&XGBoost & 56.80 & 6 \\ \cline{3-5}
    &&LightGBM & 41.25 &6 \\ \cline{3-5}
    &&DAE & 20.38 & 2\\\cline{3-5}
    &&GAT & 74.88 & 10\\\cline{3-5}
    &&GCN & 99.99 & 18\\\hline
\end{tabular}
  \vspace{-0.1in}
\end{table}


The experiment aims to test our initial hypothesis that using a graph structure to describe a performance dataset improves a predictive regression model's effectiveness. Figure~\ref{fig:hpc_mse} summarizes the performance of HPC datasets. 

From Figures~\ref{fig:hpc_mse} and~\ref{fig:ml_mse}, we can observe that: (1) The graph-based representation learning approaches outperform DNN. 
Although, DNN models can implicitly leverage higher-order correlations through connections across multiple layers; our experiments demonstrate that \textit{explicitly modeling} the relations between data samples in \gls{ping} improves the effectiveness of a predictive regression model.
(2) In Figures~\ref{fig:hpc_mse} and~\ref{fig:ml_mse}, we observe that one of our proposed methods (SSGNN or SSBGNN) outperforms GCN in all experiments across all datasets. This is because, our proposed method (GraphSAGE with self-supervision) captures information across the entire tabular dataset by allowing messages to pass across neighbors. In contrast, GCN aggregates information across neighbors of a fixed distance, thus failing to capture long-distance relationships among the samples. 
(3) Our proposed method 
performs on per or better than GAT's attention-based aggregation strategy in 36 out of 78 experiments. 
(4) Figure~\ref{fig:hpc_mse} shows that the graph-based methods outperform XGBoost and LightGBM where the target values have a wide range of distribution, such as in the case of the Vulcan dataset compared to the Catalyst dataset, as shown in Table~\ref{tab:datasets}'s mean and standard deviation columns. 
Figure~\ref{fig:ml_mse}a shows that the \gls{ssgnn} method outperforms the gradient- and neural-network-based methods in the IoT Telemetry dataset. 


 

\subsection{Impact of Missing Data}
\begin{figure}[t]
    \centering
    \includegraphics[width=\columnwidth]{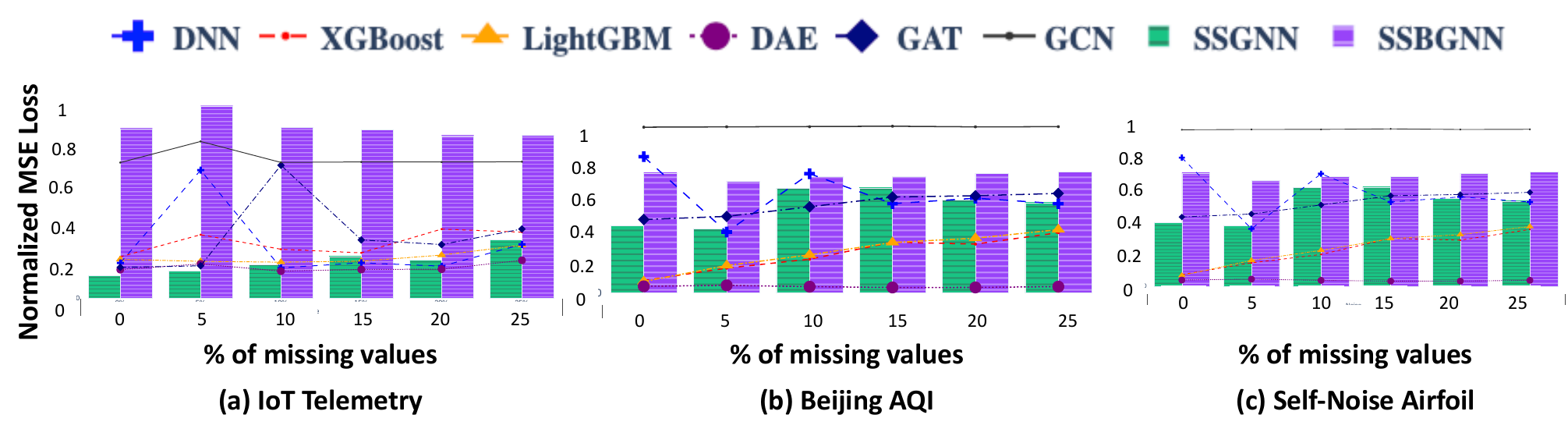}
    \caption{Normalized MSE Loss for ML datasets. The X-axis represents the \% of missing feature values and the Y-axis shows the normalized MSE for different methods (the lower, the better).}
    \label{fig:ml_mse}
    \vspace{-0.2in}
\end{figure}

This experiment aims to evaluate the impact of the \gls{ping} formulation on the performance of regression tasks when missing data is present in the features. The rationale for conducting this experiment is that streaming \gls{hpc} performance monitoring systems often miss recording feature values or cannot measure target values (unlabeled) until the application ends. To study the impact of missing data in the features on the effectiveness of graph-based representation learning techniques, we randomly inject missing values (NaN) in 5\%, 10\%, 15\%, 20\%, and 25\% of the samples and compare the generated embeddings from all the methods listed in Table~\ref{tab:result_table}. 
The GCN and GAT methods use the graph constructed using our proposed single graph construction method (Section~\ref{sec:construction}). 

From Figures~\ref{fig:hpc_mse} and~\ref{fig:ml_mse}, we observe that although the decision-tree and gradient-based methods outperform graph-based methods, our method of using GNN with self-supervision improves the efficacy of neural network-based methods in building effective representations, which enhances the performance of downstream regressions.

\subsection{Discussion }
Table~\ref{tab:result_table} summarizes the best \%-improvement in MSE loss of our proposed methods (either \gls{ssgnn} or \gls{ssbgnn}) compared to all other baseline methods. \tzi{The rationale for reporting the best of these two methods is that any of these methods being better supports our hypothesis that transforming performance data into a graph improves the performance of downstream analyses.}
The table also reports how frequently our proposed methods perform better than the baselines. From Table~\ref{tab:result_table}, we can observe that even though GAT and GCN work on the transformed graph, 
our proposed embedding learning algorithm outperforms GCN and GAT in 100\% and $>43\%$ of the experiments, respectively, for the HPC datasets. Our proposed method outperforms DNN more than $50\%$ of the times, and achieves 61.67\% improvement in MSE across all HPC datasets. 
\tzi{It is important to note that even though the tabular methods perform better for most of the HPC datasets, these methods require model rebuilding when new unlabled performance samples arrive in a streaming performance monitoring system, which makes tabular methods computationally infeasible. On the other hand, finding a neighborhood for new samples based on similarity in a graph enables quick approximation of their labels, making a graph-based performance representation suitable for streaming scenarios.}

Our experimental results show that the embedding learning pipeline with single-graph construction approach, \gls{ssgnn}, performs better than the batched-graph construction approach in nearly all cases, making \gls{ssgnn} a superior graph construction method for the \gls{ping} formulation. 

\section{Related Work}
\label{sec:related-work}
\subsection{Deep Modeling Explainability using Graphs }
Researchers have transformed tabular data into graphs to produce an explainable model called TableGraphNet~\cite{TableGraphNet}. TableGraphNet architecture builds one or more graphs for each data sample. The graph node comprises the feature attributes, and the graph edges contain the distance between the feature attributes. Using the generated graphs for the input samples, they extract node and attribute-centric features for every attribute. 
Their evaluations demonstrate that TableGraphNet architecture performs similarly to a regular Deep Neural Network model based on three classifications and eight regression datasets, 
whereas our architecture demonstrates significant improvement compared to DNN. 

\subsection{Tabular Data Prediction using Multiplex Graphs }
TabGNN~\cite{guo2021tabgnn} framework takes tabular data as input, constructs a directed multiplex graph based on the table columns referred to as features, and encodes each sample for an initial node embedding. It uses a graph neural network to produce latent feature embedding for each data sample. After getting the final representation embedding, TabGNN uses Auto Feature Engine (AutoFE) to choose significant features and then uses a Multi-Layer Perceptron to get the final prediction. Researchers observe that the framework heavily relies on AutoFE and DeepFM~\cite{guo2017deepfm}, two popular DNN-based feature engineering methods for tabular data. 
Their experiments show that utilizing AutoFE+TabGNN and DeepFM+TabGNN outperform using AutoFE or TabGNN alone, but requires longer time to train. In contrast, our method leverages a similar graph-building method without the expensive DNN-based feature engineering step, thus requiring less training and testing time. As TabGNN uses DNN based approaches and our methods outperform DNN based approaches in most of the settings, we anticipate that our methods will either perform on-per or better than TabGNN. 

\subsection{Knowledge Graph for Efficient Meta-learning }
Recent work such as Automated Relational Meta-learning (ARML)~\cite{yao2020automated} shows that graph structures improve the performance of Meta-Learning~\cite{vilalta2002perspective}. 
On a 2D toy application with random numbers, ARML leverages the meta-knowledge graph to obtain a more fine-grained structure than other gradient-based meta-learning implementations. The single limitation of this implementation is that it can only be used with meta-learning algorithms and has been evaluated on image classification datasets. The ARML architecture uses DNN to transform the input data into a feature representation. This architecture complements our approach, where our proposed method can convert tabular data to a graph formulation and then leverage ARML to produce a fine-grained representation.

\section{Conclusions}
\label{sec:conclusions}
This paper proposes a new data transformation approach that converts tabular data into graphs and develops a new representation learning pipeline using a self-supervision-based automated edge-inference-based technique. Graph representation learning leverages relationships between samples and features to create fine-grained embeddings, effectively improving the model's performance and outperforming the standard deep learning-based techniques.
We evaluate our approach using different models on multiple HPC and ML datasets and find that relationships captured between samples and features in a graph formulation make regression tasks robust against missing values and perform better. 

\section{Acknowledgement}
The work was supported in part by the grant DE-SC0022843 funded by the U.S. Department of Energy, Office of Science. The work was also supported in part by the Exascale Computing Project (ECP) funded by the U.S. Department of Energy, Office of Science. A part of the research was also supported by the Research Enhancement Program at Texas State University.

\bibliographystyle{IEEEtran}
\bibliography{main,tzi,proposal}

\end{document}